\documentclass{article}
\usepackage{spconf,amsmath, graphicx, amssymb, amsfonts}
\usepackage{caption}
\usepackage{cite}
\usepackage{algorithmic}
\usepackage{textcomp}
\usepackage{comment}
\usepackage{color}
\usepackage{multirow}
\usepackage{array}
\usepackage{xcolor}
\usepackage{hyperref}
\usepackage{url}

\hypersetup{
    colorlinks=true,
    linkcolor=red,
    filecolor=magenta,      
    urlcolor=blue,
    citecolor=red,
    }

\title{HRTR: A Single-Stage Transformer for Fine-Grained Sub-Second Action Segmentation in Stroke Rehabilitation}

\vspace{-0.5cm}

\name{
Halil Ismail Helvaci\textsuperscript{\,1} \enspace
Justin Philip Huber\textsuperscript{\,2} \enspace
Jihye Bae\textsuperscript{\,1} \enspace
Sen-ching Samson Cheung\textsuperscript{\,1,3} 
\thanks{Research reported in this publication was supported by the Igniting Research Collaborations (IRC) at the University of Kentucky.}}
\address{
\textsuperscript{\,1}Department of Electrical and Computer Engineering, University of Kentucky, Lexington, KY, USA\\
\textsuperscript{\,2}College of Medicine, University of Kentucky, Lexington, KY, USA\\ 
\textsuperscript{\,3}Department of Electrical and Computer Engineering, University of California, Davis, CA, US \enspace \\
{\tt\small halil.helvaci@uky.edu \enspace
\tt\small  justin.huber@uky.edu \enspace
\tt\small  jihye.bae@uky.edu \enspace
\tt\small  sccheung@ieee.org \enspace
}
}

\begin{document}
\maketitle
\begin{abstract}

Stroke rehabilitation often demands precise tracking of patient movements to monitor progress, with complexities of rehabilitation exercises presenting two critical challenges: fine-grained and sub-second (under one-second) action detection. In this work, we propose the High-Resolution Temporal Transformer (HRTR), to time-localize and classify high-resolution (fine-grained), sub-second actions in a single-stage transformer, eliminating the need for multi-stage methods and post-processing. Without any refinements, HRTR outperforms state-of-the-art systems on both stroke related and general datasets, achieving Edit Score (ES) of 70.1 on StrokeRehab Video, 69.4 on StrokeRehab IMU, and 88.4 on 50Salads. 

\end{abstract}


\section{INTRODUCTION}

Stroke is a leading cause of disability, affecting over 795,000 individuals annually in the United States. Among stroke survivors, 77.4\% experience arm impairments, which significantly hinder their ability to perform daily activities independently and diminish their quality of life \cite{tsao2023heart, lawrence2001estimates}. Rehabilitation focused on arm movements plays a crucial role in addressing these limitations, enabling patients to regain functional independence and reducing the burden on caregivers. Assessments of the patient are crucial to the cycle of rehabilitation -- a cycle that involves identifying needs, implementing interventions, and monitoring progress \cite{world2011world}. Observation-based assessments have long been the standard in clinical practice; however, they have well-documented limitations, including difficulty detecting subtle changes \cite{murphy2011kinematic} and weak correlations with real-world outcomes measured by sensor-based activity monitors \cite{waddell2017does}. They also face challenges identifying purposeful movements and nuanced variations of fine motor movements \cite{chen2021novel}. A need exists for more precise assessment of the post-stroke upper rehabilitation.

A key enabler of precise assessment is temporal action segmentation, which identifies and classifies movements from continuous streams of sensor or video data. However, the complexities of rehabilitation exercises present two critical challenges: fine-grained action detection and sub-second action duration. Fine-grained actions involve subtle and nuanced movements, such as differentiating between reaching and stabilizing motions, which are essential for assessing progress in therapy. Meanwhile, sub-second actions are short in duration, often less than a second, demanding high temporal resolution for accurate detection. These challenges are particularly relevant in stroke rehabilitation, where even the smallest movements can hold significant clinical importance, underscoring the need for advanced segmentation techniques.

To address these challenges, the StrokeRehab dataset provides a high-resolution resource specifically tailored for stroke rehabilitation research \cite{kaku2022strokerehab}. It captures fine-grained, sub-second actions using a multi-modal setup, including inertial measurement units (IMUs) and video cameras. Traditional approaches, such as recurrent neural networks (RNNs) and convolutional neural networks (CNNs), struggle with long-range dependencies and variable-length actions \cite{sun2017lattice, farha2019ms}. Transformer-based models, while more effective in capturing global context, frequently rely on complex multi-stage frameworks or computationally intensive post-processing steps \cite{yi2021asformer, wang2024efficient, liu2023diffusion, van2023aspnet}, limiting their efficiency and applicability in medical settings.

In this study, we introduce HRTR, a single-stage transformer model for sub-second action segmentation in stroke rehabilitation. Despite its simplicity, HRTR outperforms state-of-the-art models through careful design. The model projects feature vectors into embeddings and incorporates temporal positional information via sinusoidal encoding. A sliding window approach is employed to efficiently handle long sequences, enabling the model to capture fine-grained actions while preserving global context. By combining efficient temporal encoding with sliding window processing, HRTR captures fine-grained temporal details more effectively than existing methods. Beyond addressing the specific demands of stroke rehabilitation, HRTR also demonstrates strong generalization to diverse action event datasets such as 50Salads. Our model establishes a new state of the art on the StrokeRehab Video and IMU datasets, as well as 50Salads, surpassing previous works, showcasing its effectiveness and potential for broader applications in action segmentation. This work contributes to the field by:

1. introducing a single-stage transformer model that effectively captures fine-grained, sub-second actions, addressing the specific challenges of stroke rehabilitation, and

2. demonstrating the model's generalization capability across diverse datasets, such as 50Salads, and establishing a new state of the art on the StrokeRehab Video and IMU datasets.

\vspace*{-0.4cm}
\section{RELATED WORK}

Action segmentation has traditionally relied on fixed-duration temporal models, which often struggled to capture long-range dependencies and complex temporal dynamics. Recurrent Neural Networks (RNNs), particularly Long Short-Term Memory (LSTM) networks \cite{sun2017lattice}, improved temporal modeling but encountered challenges with vanishing gradients in long sequences \cite{vaswani2017attention}. Temporal Convolutional Networks (TCNs), such as the Multi-Stage Temporal Convolutional Network (MS-TCN) \cite{farha2019ms}, addressed these limitations by effectively modeling long-range dependencies but frequently suffered from over-segmentation. Techniques like the Action Segmentation Refinement Framework (ASRF) \cite{ishikawa2021alleviating} introduced boundary refinement to address this issue.

Transformer models, known for their ability to model global temporal dependencies, have shown promise in action segmentation. ASFormer \cite{yi2021asformer} introduced a two-stage encoder-decoder transformer architecture for action segmentation. Baformer \cite{yi2021asformer} proposed a transformer-based model that incorporates both local and global context for enhanced action recognition. ASPNet \cite{van2023aspnet} presented a novel approach using action-specific attention to focus on the most relevant parts of the input sequence. Transformers have further found applications in specialized domains, such as autism-related behavior prediction \cite{helvaci2024localizing} and relevance detection in surgical videos \cite{ghamsarian2021relevance}, demonstrating the versatility of these models in diverse contexts. Recently, DiffAct \cite{liu2023diffusion} leveraged diffusion models to refine action boundaries and generate smooth predictions by iteratively de-noising noisy action sequences. Moreover, post-processing techniques, such as ASRF’s boundary prediction \cite{ishikawa2021alleviating} and UARL’s uncertainty learning \cite{chen2022uncertainty}, further enhanced segmentation accuracy.

Despite advancements, fine-grained segmentation of sub-second actions remains a challenge. Datasets like StrokeRehab \cite{kaku2022strokerehab} exemplify the need for models that capture subtle, short-duration movements. While an LSTM-based sequence-to-sequence model in \cite{kaku2022strokerehab} captured fine-grained dynamics, it required a multi-stage structure and post-processing to remove duplicated predictions, limiting scalability and efficiency. Similarly, existing transformer-based methods employ multi-stage structures. The hierarchical architectures can lead to information loss during feature aggregation, while self-attention mechanisms optimized for broad temporal contexts may struggle to detect transient, rapidly occurring actions. This can lead to over-smoothed outputs, where temporal details are excessively averaged, blurring distinct action boundaries and resulting in imprecise boundary predictions. To address these limitations, we propose HRTR, a single-stage transformer model that captures global dependencies and fine-grained temporal patterns without complex architectures or extensive post-processing. HRTR employs a sliding window approach to focus attention on shorter temporal scales while maintaining global context through overlapping windows, enabling robust sub-second segmentation in specialized datasets with subtle, high-frequency movement variations like StrokeRehab.

\section{DATASET}

\subsection{StrokeRehab}

The StrokeRehab dataset, developed by Kaku et al. \cite{kaku2022strokerehab}, includes 3,372 trials from 51 stroke-impaired patients and 20 healthy subjects, designed for stroke rehabilitation research. It contains 120,891 annotated functional primitives across nine activities, such as feeding and brushing teeth. The annotations, labeled by trained coders under expert supervision, have high inter-rater reliability with Cohen's kappa $\geq$ 0.96. 

\begin{figure}[!t]
    \centering
    \small 
    \begin{tabular}{m{1.1cm}|m{2.8cm}|m{2.8cm}}
    \hline
    Activity & Description & Example  \\ \hline
    
    Rest      & A stationary state where there is no significant movement.   & Sitting still with hands resting on a table.     \\ \hline
    Reach     & Extending an arm toward a target.    &  Stretching the arm to pick up the glass or bottle.    \\ \hline
    Retract   & Pulling an arm back after reaching or performing an action.   & Bringing the arm back after placing the glass or bottle.\\ \hline
    Stabilize & Holding the target object steady to maintain control.     &  Holding the bottle to allow the other hand to open the cap.   \\ \hline
    Transport & Moving a target from one location to another. & Moving the bottle to pour some water into the glass.          \\ \hline
    \end{tabular}
    \includegraphics[width= 0.9\linewidth]{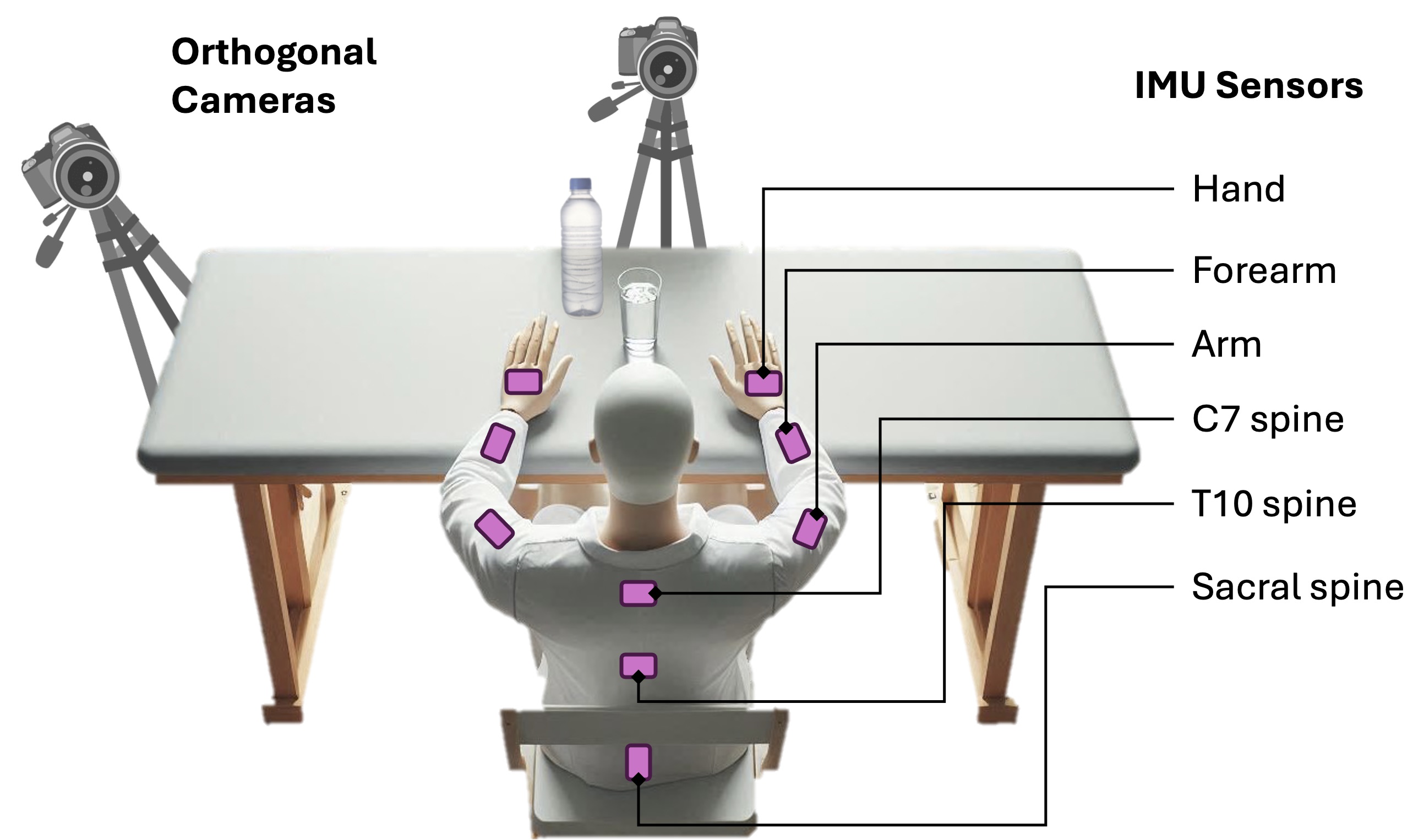}
    \captionsetup{justification=centering}
    \caption{Table of primitive motions (top) and data capture setup (bottom) with 2 orthogonal cameras and 9 IMUs on arms, hands, forearms, and back.}
    \label{fig:definitions_and_IMUplacement}
\end{figure}

Nine IMUs recorded upper body motion at C7, T12, pelvis, arms, forearms, and hands, capturing 76 kinematic data points at 100 Hz, shown in Fig. \ref{fig:definitions_and_IMUplacement}. These included joint angles, 3D quaternions, and accelerations. Video was captured by two cameras positioned orthogonally, at 1088 x 704 resolution and 60 or 100 fps, depending on the trial. Since the raw videos were withheld for privacy reasons, features were extracted by \cite{kaku2022strokerehab} as described in Section \ref{sec:feature extraction}. The dataset is accessible on SimTK: \url{https://simtk.org/projects/primseq}.

\vspace*{-0.2cm}
\subsection{50Salads}
The 50Salads dataset \cite{stein2013combining} consists of 50 videos featuring 17 action classes related to salad preparation, utilized for action segmentation and detection. On average, each video is 6.4 minutes and includes 20 action instances. The tasks were performed by 25 subjects, with each subject preparing two distinct salads.

\vspace*{-0.2cm}
\subsection{Video Feature Extraction}
\label{sec:feature extraction}

High-dimensional video data often contains redundant or irrelevant information, complicating action segmentation. Feature extraction reduces data complexity. 
For the StrokeRehab dataset, we use the pre-extracted video features provided by \cite{kaku2022strokerehab}, obtained using the X3D model \cite{feichtenhofer2020x3d} pre-trained on Kinetics \cite{kay2017kinetics} to capture spatiotemporal features like motion dynamics, spatial patterns, and temporal progression. The authors \cite{kaku2022strokerehab} fine-tuned the model on the StrokeRehab dataset to address the action disparity between high-level actions in Kinetics (running, climbing, etc.) and the fine-grained actions in StrokeRehab (reach, transport, etc.). For 50Salads , we use pre-extracted features obtained from the I3D \cite{carreira2017quo} model, which was trained on the Kinetics \cite{kay2017kinetics} dataset, following the approach of previous works.

\vspace*{-0.3cm}
\section{Action Segmentation}

Given a sequence of input features \textbf{X} $ = \{ x_1, x_2, \ldots, x_T\}$, where $T$ denotes the total number of discrete time steps, the goal of 
action segmentation is to generate the corresponding sequence of action labels, \textbf{Y} $ = \{y_1, y_2, ..., y_T\}$.  This section describes the details of our proposed HRTR system in solving the action segmentation problem.

\subsection{Model Architecture}

\begin{figure}[ht]
    \centering
    \includegraphics[width= 0.9\linewidth]{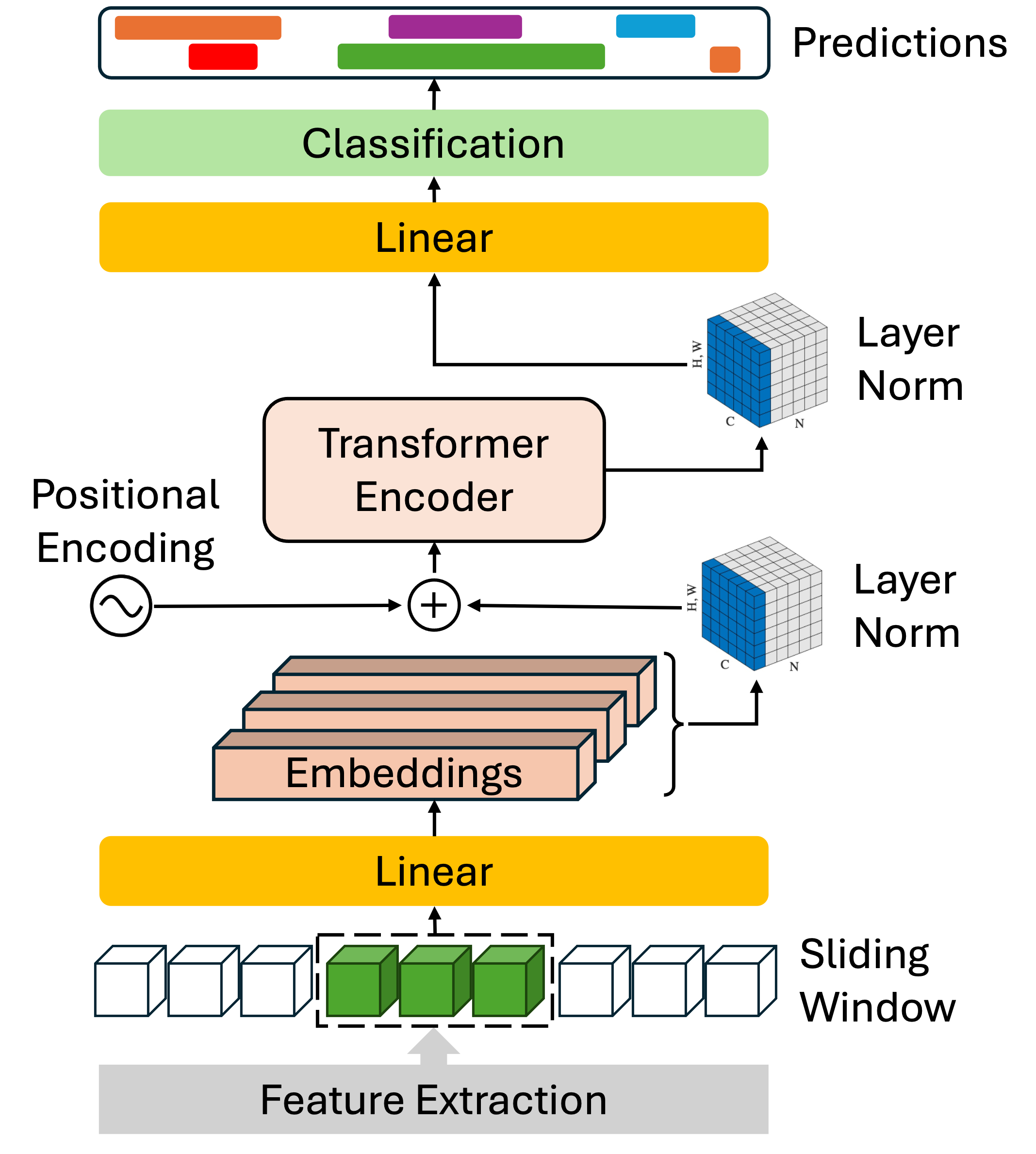}
    \captionsetup{justification=centering}
    \caption{Visualization of the HRTR action prediction pipeline}
    \label{fig:model}
\end{figure}

HRTR is a single-stage transformer encoder designed for sub-second action segmentation. It is designed with robust regularization mechanisms that enhance its performance on high-resolution temporal datasets. The model architecture is depicted in Fig. \ref{fig:model}. We employ a sliding window approach, segmenting input sequences into overlapping windows of $w=200$ for the video data, 500 for the IMU data and 5000 for the 50Salads dataset. The stride is $s=10$ for both IMU and video datasets, and 500 for the 50Salads dataset. $w$ and $s$ are hyper-parameters tuned to match the action events of interest. 

Input features are projected into 1024-dimensional embeddings using a linear layer with dropout and GELU activation, followed by layer normalization. Temporal information is incorporated into the extracted embeddings using the positional encoding approach 
from \cite{vaswani2017attention}. The encoded embeddings are subsequently processed by a multi-layer Transformer encoder with 3 encoder layers, 4 attention heads, and hidden model dimension of 512, followed by layer normalization. The output is then passed through a linear layer and a classification head to generate action class probabilities for each time step. Both IMU and video models share this structure. As we use a different feature embedding for the 50Salads dataset, the Transformer encoder is modified to include 3 encoder layers, 2 attention heads, and a hidden model dimension of 256, while maintaining the same 1024-dimensional final linear layer. The training configurations, including batch size, learning rate schedules, and optimization strategies, are detailed in Section \ref{sec:experimental_setup}.

To tackle the class imbalance present in the StrokRehab and 50Salads datasets, we apply focal loss \cite{tian2022focal}, which adjusts cross-entropy loss based on confidence, mitigating class dominance: $L_{focal}(p_t) = -\alpha(1 - p_t)^\gamma log(p_t)$ where $p_t = \text{softmax}(z)$, where $z$ denotes the logits, $\alpha$ is the class weighting factor and $\gamma \geq 0$ is the focus parameter.

By integrating efficient temporal encoding with a sliding window approach, the model effectively captures fine-grained temporal details. The sliding-window strategy allows the model to learn local temporal patterns within each window while maintaining global context through overlapping regions. This overlap ensures smooth transitions between windows, reducing boundary effects that could negatively impact prediction accuracy. Moreover, the approach mitigates memory constraints, facilitates the detection of short-duration actions, and minimizes information loss over long sequences.


\section{Experiments}

\subsection{Evaluation Metrics}
\label{sec:metrics}


Segmentation performance is evaluated using the Levenshtein distance \cite{kaku2022strokerehab}, which computes the minimum number of insertions, deletions, and substitutions required to transform the predicted sequence $P$ into the ground-truth sequence $G$, denoted as $L(G,P)$. For instance, $G = [reach, idle, retract]$ and $P = [reach, stabilize]$ yields $L(G,P) = 2$ (one substitution and one insertion). The edit score \cite{farha2019ms, yi2021asformer, ishikawa2021alleviating, kaku2022strokerehab} normalizes this distance: $\text{ES}(G,P) = 1 - \frac{L(G,P)}{\text{max}(|G|,|P|)} \times 100$. To address the leniency in normalizing by the maximum sequence length, we also use the Action Error Rate (AER) \cite{kaku2022strokerehab}: $\text{AER}(G,P) = \frac{L(G,P)}{len(G)}$. We also use standard classification metrics including Sensitivity: recall of positive cases, Specificity: recall of negative cases, and the F1 score: the harmonic mean of precision and recall.

\subsection{Experimental Setup}
\label{sec:experimental_setup}

The StrokeRehab IMU and video models, along with the 50Salads model, are trained using a consistent experimental setup with dataset-specific adjustments. For StrokeRehab, the models are trained for 25 epochs with a batch size of 8 and an initial learning rate of $10^{-3}$, which is reduced by a factor of 0.01 if the focal loss does not improve over 5 consecutive epochs. In contrast, the 50Salads model is trained for 10 epochs with a batch size of 2. A dropout rate of 0.2 is applied during training across both datasets to mitigate overfitting. We follow the same train/test splits as previous methods \cite{kaku2022strokerehab, yi2021asformer}.

Model optimization is performed using Stochastic Gradient Descent (SGD) with a momentum of 0.9 and a weight decay of $10^{-4}$. The focal loss parameters are set to $\alpha = 25$ and $\gamma = 2$ for both models. Gradient clipping is employed to enhance training stability, with a maximum norm of 5 for StrokeRehab and 60 for 50Salads. During inference, all models process sequences in non-overlapping windows. All experiments are conducted on an NVIDIA RTX A6000 GPU.

\subsection{Comparison with State-of-the-Art Methods}

The proposed models are evaluated against previous benchmarks on all the datasets and the results aresummarized in Table \ref{tab:model_comparison}. Evaluation metrics include Edit Score (ES, higher is better) and Action Error Rate (AER, lower is better) as defined in Section \ref{sec:metrics}. Models marked with $+$ denote variants enhanced with smoothing windows, which are applied to reduce prediction noise by averaging model outputs over a defined temporal window. We used a smoothing window size of 25 for the StrokeRehab datasets and 200 for the 50 Salads dataset. Models marked with an asterisk (*) were selected based on the best validation frame-wise accuracy.

\begin{table}[ht]
\caption{Comparison with the state-of-the-art on StrokeRehab Video, IMU and 50Salads}
\centering
\Large
\resizebox{\columnwidth}{!}{
\begin{tabular}{l|c|c|c|c|c|c|c}
\hline
\multirow{2}{*}{Model} &\multirow{2}{*}{Venue} & \multicolumn{2}{c|}{\textbf{Video}} & \multicolumn{2}{c|}{\textbf{IMU}} & \multicolumn{2}{c}{\textbf{50 Salads}}  \\ \cline{3-8}
 &  & ES $\uparrow$  & AER $\downarrow$ & ES $\uparrow$  & AER $\downarrow$  & ES $\uparrow$  & AER $\downarrow$  \\
\hline
MS-TCN* \cite{farha2019ms}  & CVPR 2019 & 60.7 & 0.408 & 66.9 & 0.372 &68.8  &0.47  \\
MS-TCN \cite{farha2019ms}   & CVPR 2019 & 62.2 & 0.392 & 68.9 & 0.330 &70.8  &0.43  \\
MS-TCN$^{+}$ \cite{farha2019ms}                 & CVPR 2019 & 62.7 & 0.390 & 68.8 & 0.317 &76.4  &0.32  \\
ASRF* \cite{ishikawa2021alleviating}  & CVPR 2021 & 56.9 & 0.449 & 68.2 & 0.328  &74.0  &0.34  \\
ASRF \cite{ishikawa2021alleviating}   & CVPR 2021 & 58.7 & 0.436 & 67.9 & 0.349 &75.2  &0.33   \\
Seg2Seq \cite{kaku2022strokerehab}    & NeurIPS 2022 & 67.6 & 0.322 & 63.0 & 0.337  &76.9  &0.30  \\
Raw2Seq \cite{kaku2022strokerehab}    & NeurIPS 2022 & 66.6 & 0.329 & 68.8 & \textbf{0.305} &69.4  &0.54   \\ 
ASFormer \cite{yi2021asformer}        &  BMVC 2021 & - & - & - & - &79.6  & - \\
DiffAct \cite{liu2023diffusion}       & ICCV 2023 & - & - & - & - &85.0   & - \\
ASPnet \cite{van2023aspnet}           & CVPR 2023 & - & - & - & - &87.5  & - \\
BaFormer \cite{wang2024efficient}     & RS 2024 & - & - & - & - &84.2  & - \\ \cline{1-8}
HRTR (ours)                &  & 69.8 & 0.302 & 68.9 & 0.311 &85.1  &0.149   \\ 
HRTR$^{+}$ (ours)    &  & \textbf{70.1} & \textbf{0.299} & \textbf{69.4} & 0.306 &\textbf{88.4}  &\textbf{0.116}  \\ 
\hline
\end{tabular}}
\label{tab:model_comparison}
\end{table}

In StrokRehab video, HRTR achieved an ES of 69.8 and an AER of 0.302, while HRTR$+$ outperformed all competing methods with an ES of 70.1 and an AER of 0.299, establishing a new state-of-the-art performance. For the IMU modality, HRTR attained an ES of 68.9 and an AER of 0.311, with HRTR$+$ further improving these results to an ES of 69.4 and an AER of 0.306. Notably, Raw2Seq achieved a marginally lower AER of 0.305 in this modality, slightly outperforming our model. On the 50Salads dataset, HRTR achieved an ES of 85.1 and an AER of 0.149, while HRTR$+$ demonstrated significant improvements, achieving an ES of 88.4 and an AER of 0.116. This performance exceeds the previous best result by ASPnet, which achieved an ES of 87.5. The results underscore the efficacy of our proposed models, particularly in the 50Salads modality, where HRTR$+$ sets a new benchmark for action segmentation tasks.

\begin{figure}[ht]
    \centering
    \includegraphics[width= 1.0\linewidth]{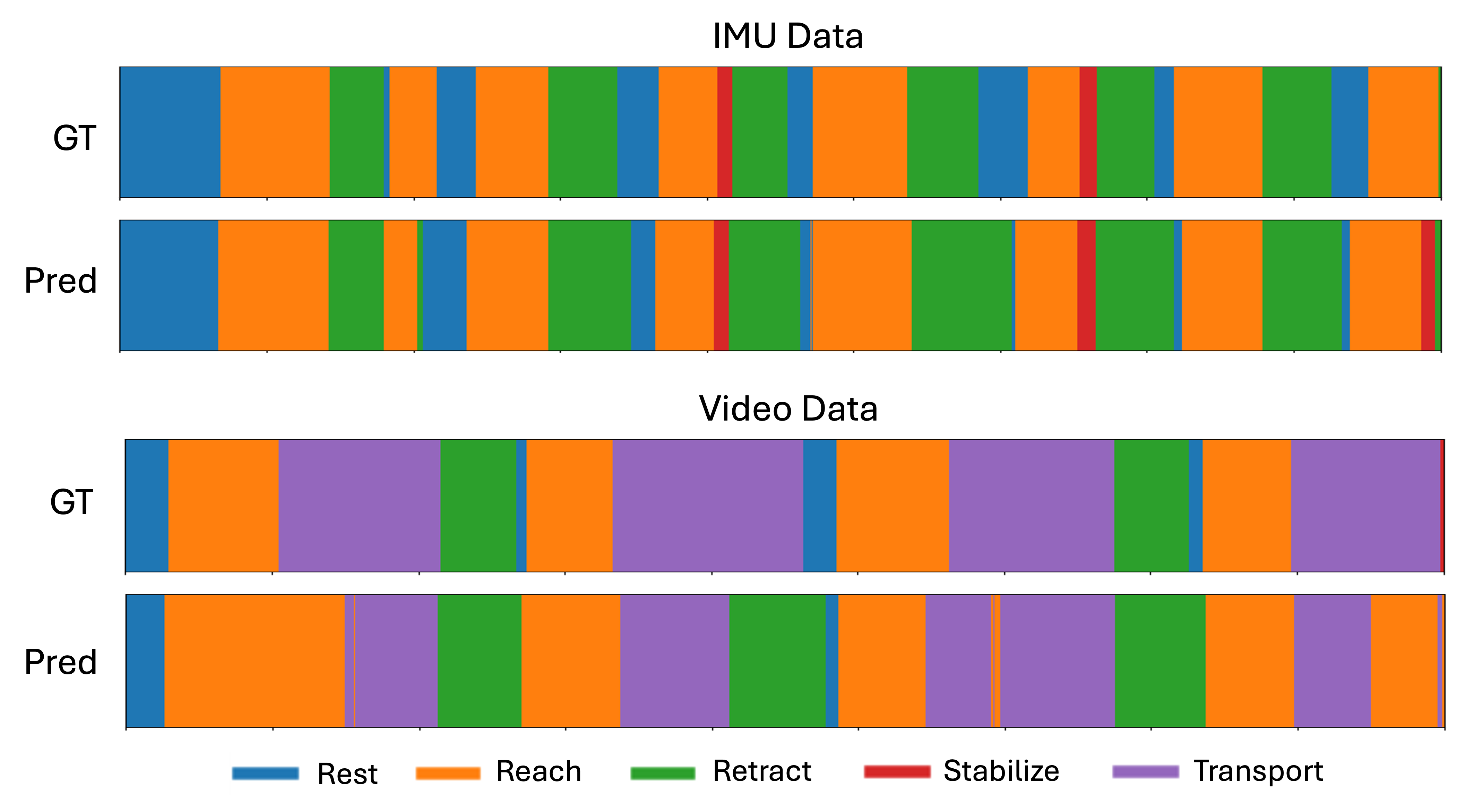}
    \captionsetup{justification=centering}
    \caption{Visualization of action predictions (Pred) vs. ground truth (GT) labels on IMU Data (top) and Video Data (bottom)}
    \label{fig:combined_predictions}
\end{figure}

Fig.~\ref{fig:combined_predictions} compares HRTR’s predictions (Pred) with ground truth (GT) on two randomly chosen sequences from the IMU (top) and video test sets respectively. While predictions align well overall, the IMU results show occasional over-segmentation, especially between \emph{Rest} and \emph{Reach} likely due to sensor fluctuations. In video data, under-segmentation occurs, notably between \emph{Transport} and \emph{Reach}. This may be due to the model’s sensitivity to overlapping visual features when consecutive actions exhibit similar motion patterns. These results indicate strong temporal modeling with room for improvement in boundary localization.

\vspace*{-0.4cm}
\subsection{Classification Performance}

\begin{figure}[ht]
    \centering
    \includegraphics[width= 0.8\linewidth]{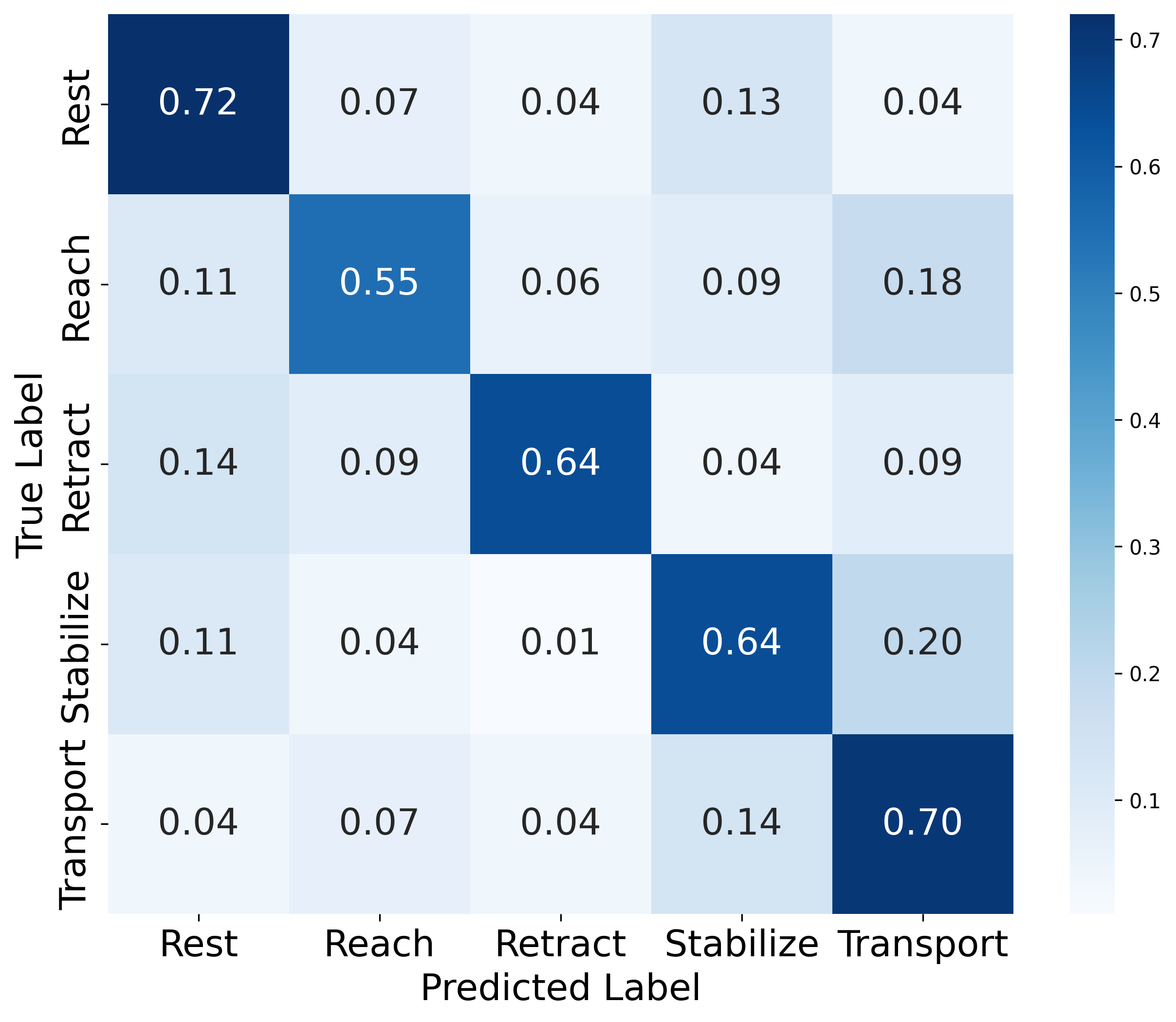}
    \captionsetup{justification=centering}
    \caption{Visualization of the confusion matrix for the StrokeRehab video dataset}
    \label{fig:video_confusion_matrix}
\end{figure}

Table \ref{tab:cls_combined} summarizes the classification performance for primitive actions on StrokeRehab. IMU data excels in capturing fine-grained temporal motion patterns, resulting in higher sensitivity and F1-scores for dynamic actions like \emph{Reach} and \emph{Retract}, while video data leverages spatial context to achieve higher specificity, reducing false positives. The confusion matrix for the video model in Fig.~\ref{fig:video_confusion_matrix} reveals strong performance for actions like \emph{Transport} and \emph{Retract}, but challenges arise in distinguishing actions with overlapping motion characteristics, such as \emph{Reach} and \emph{Stabilize}. These findings suggest that combining IMU and video data through multimodal fusion could enhance classification accuracy by leveraging temporal precision and spatial context, ultimately improving the reliability of stroke rehabilitation monitoring systems.

\begin{table}[!h]
\caption{Evaluation of action classification across the StrokeRehab Video and IMU datasets}
\centering
\small
\resizebox{\columnwidth}{!}{
    \begin{tabular}{l|c c c|c c c} 
    \hline
         \multirow{2}{*}{Action}   & \multicolumn{3}{c|}{Video} & \multicolumn{3}{c}{IMU} \\ \cline{2-7}
              & Sens. & Spec. & F1 & Sens. & Spec. & F1 \\ \hline
    Rest      & 0.70        & 0.92        & 0.66     & 0.69        & 0.93        & 0.67     \\ 
    Reach     & 0.52        & 0.96        & 0.59     & 0.59        & 0.94        & 0.64     \\ 
    Retract   & 0.60        & 0.97        & 0.62     & 0.64        & 0.97        & 0.69     \\ 
    Stabilize & 0.59        & 0.91        & 0.63     & 0.65        & 0.90        & 0.60     \\ 
    Transport & 0.77        & 0.85        & 0.70     & 0.72        & 0.88        & 0.71     \\ \hline
    
    \end{tabular}
}
\label{tab:cls_combined}
\end{table}

\vspace*{-0.4cm}
\section{ABLATION STUDY}

An ablation study was conducted to investigate the impact of window size on HRTR$+$ performance using the StrokeRehab dataset, detailed in Table~\ref{table:window_ablation}. The study evaluated various window size $w$, ranging from 100 to 1500, with stride $s$ selected based on the best-performing ranges observed during preliminary experiments. Smaller windows, 200 and 500, improved video performance (best at 200: ES 70.1, AER 0.299), while a window size of 500 yielded the best IMU results (ES 69.4, AER 0.306). These findings highlight the importance of tailoring window sizes to the specific characteristics of the input modality. We hypothesize that video data requires finer granularity for rich spatial-temporal details, whereas IMU data benefits from larger windows that capture broader temporal context reducing sensitivity to noise. A detailed analysis of stride effects will be explored in future work.

\begin{table}[ht]
\caption{Ablation study on the impact of window size on the
StrokeRehab dataset}
\centering
\small
\resizebox{\columnwidth}{!}{
\begin{tabular}{l|c|c|c|c|c|c}
\hline
\multicolumn{7}{c}{StrokeRehab} \\ \hline
\multirow{2}{*}{Model} & \multicolumn{2}{c|}{Video} & \multicolumn{2}{c|}{IMU} & \multirow{2}{*}{Win. Size} & \multirow{2}{*}{Stride}      \\ \cline{2-5}
                       & ES $\uparrow$  & AER $\downarrow$  & ES $\uparrow$ & AER $\downarrow$ & & \\ \hline

HRTR$^{+}$ (ours)     &62.0  &0.388     &67.6    &0.324  &1500    &500  \\
HRTR$^{+}$ (ours)    &66.8  &0.332     &68.9  &0.314  &1000  &500\\
HRTR$^{+}$ (ours)     &68.2  &0.318     &66.9    &0.331  &800    &500  \\
HRTR$^{+}$ (ours)     &68.7   &0.313      &\textbf{69.4}    &\textbf{0.306}  &500    &10  \\
HRTR$^{+}$ (ours) &\textbf{70.1}  &\textbf{0.299} &66.8    &0.332  &200    &10  \\
HRTR$^{+}$ (ours)    &68.0  &0.320    &63.8    &0.364  &100    &10  \\

\hline
\end{tabular}}
\label{table:window_ablation}
\end{table}

\vspace*{-0.5cm}
\section{DISCUSSION AND CONCLUSION}
\label{conclusion}

In this study we presented HRTR, a single-stage transformer model for high temporal resolution action segmentation, addressing fine-grained and sub-second actions without the need for multi-stage frameworks. Evaluated on the StrokeRehab and 50Salads datasets, HRTR achieved superior performance. Its efficiency and precision establish a strong foundation for applications requiring detailed temporal action analysis, such as stroke rehabilitation, while offering a scalable approach for broader use cases. 

\bibliography{refs}
\end{document}